# Membership Function Assignment for Elements of Single OWL Ontology

Olegs Verhodubs
oleg.verhodub@inbox.lv

**Abstract.** This paper develops the idea of membership function assignment for OWL (Web Ontology Language) ontology elements in order to subsequently generate fuzzy rules from this ontology. The task of membership function assignment for OWL ontology elements had already been partially described, but this concerned the case, when several OWL ontologies of the same domain were available, and they were merged into a single ontology. The purpose of this paper is to present the way of membership function assignment for OWL ontology elements in the case, when there is the only one available ontology. Fuzzy rules, generated from the OWL ontology, are necessary for supplement of the SWES (Semantic Web Expert System) knowledge base. SWES is an expert system, which will be able to extract knowledge from OWL ontologies, found in the Web, and will serve as a universal expert for the user.

## 1. INTRODUCTION

The emergence of the Web marked a new stage in the development of mankind. From this time the information became electronically accessible wherever the Web is available. Currently billions of static documents in the Web are used by hundreds of millions of users, and the number of new documents as the number of users is constantly growing [1]. Along with the positive consequences of the rapid growth of the Web, this growth has its own drawbacks. The exponential growth of the Web makes it increasingly difficult to find, to access, to present and to maintain information of use to a wide variety of users [1].

The Semantic Web and the Semantic Web technologies offer a new approach to managing information [2]. This approach is designed to neutralize the negative manifestations associated with the rapid growth of the Web. It is assumed that the Semantic Web and the Semantic Web technologies will make it possible to organize and find information based on meaning, not just text, to improve the way information is presented and to integrate information from heterogeneous sources [2]. Ontologies are one of the key innovations of the Semantic Web, and largely the uses of ontologies promise to nullify the negative manifestations of the exponential growth of the Web. Ontologies are the basic resource for the functioning of the Semantic Web Expert System (SWES), an expert system, which is being implemented to provide expert assistance to the user according to the entered request. The implementation of the SWES is the final goal of the research [3].

The main purpose of this paper is to develop the way of membership function assignment for OWL ontology elements. The values of membership function, assigned for OWL ontology elements, are necessary for fuzzy rule generation [4]. Generated fuzzy rules are stored in the SWES knowledge base and are used by the semantic reasoner of the SWES. And although the task of membership function assignment for OWL ontology elements has already been discussed previously, this was done for the case when several OWL ontologies of one domain were available, and they were merging into a single OWL ontology [4]. In this case the process of ontology merging is combined with the process of membership function assignment for OWL ontology elements. But the situation may occur, when the only one OWL ontology of a certain domain is available. This might occur if the only one ontology was found after the user typed his request in the SWES, and after the web search engine of the SWES processed this request. In such a case there is a need of a new way of membership function assignment for OWL ontology elements, and this paper is dedicated to this task.

This paper is structured as follows. The next section explains the principles of membership function assignment for OWL ontology elements in the case, when the only one OWL ontology of a certain domain is available. The third section of this paper expounds membership function assignment for OWL ontology elements in details. Conclusions complete this paper.

## 2. PRINCIPLES OF MEMBERSHIP FUNCTION ASSIGNMENT FOR ONTOLOGY ELEMENTS

There are a lot of OWL ontologies in the Web, and this fact can be easily verified using one of the known semantic web search engines as SWOOGLE [5] or WATSON [6]. Naturally this fact is very positive thing for the SWES, because OWL ontologies are its basic source for supplementation of its knowledge base with rules, generated from these ontologies. Basically, different ontologies belong to different domains, but it is important to the SWES that several ontologies would belong to similar domain. In such a case, merging OWL ontologies of more or less belonging to the same domain and also considering repeating and missing ontology elements (hereinafter ontology elements are classes, different relations, properties and other constructs of ontology), it is possible to obtain such ontology, which belongs to the domain more, than each ontology individually. Slightly different composition of OWL ontologies, which nevertheless belong to the same domain, can serve as a basis for membership function assignment. This means that the ontology element, which occurs in ontologies more frequently, has the higher value of membership function and vice versa.

It is obvious that other basis for assignment of membership functions is necessary, when there is the only one available OWL ontology of a certain domain. The basic reason why this situation may arise is that the area of the Semantic Web is rather young, and there are no several ontologies for some domains. One more possible reason for this situation is the presence of one and exhaustive ontology for some domains. This may occur currently and in the future for very specific domains. Thus, it follows that the task of membership function assignment for ontology elements in the presence of the only one ontology should be solved in order to provide the necessary efficiency of the being developed Semantic Web Expert System.

In general OWL ontologies are not needed for the Semantic Web Expert System in themselves, whether it is the one ontology or several ontologies. Ontologies are necessary for generation of rules from them. Consequently, obtained rules or rather the inferences, derived from rule premises, are needed ultimately. So, assignment of membership functions for ontology elements in the case, when there is the only one available ontology, should be produced, based on the premises of generated rules. This means that if it is possible to generate several rules from OWL ontology, where the premises are similar, but the conclusions are different, then it is necessary to assign membership functions to these premises, which are equal to one, divided by the number of repeating premises. This can be illustrated by means of list of rules with similar premises and different inferences, which can be generated from ontology:

**IF** $X_1$ **THEN** $Y_1$, (1)

**IF** $X_1$ **THEN** $Y_2$, (2)

…………………..

**IF** $X_1$ **THEN** $Y_n$, (n)

where n is index number of generated rule.

Considering the identical premises in different rules, it is possible to assign the values of membership function (μ) to these premises, which can be calculated according to the formula:

$$\mu (X_1) = 1 / n,$$

where n is the number of rules with repeating premises.

Thus, the list of rules with the values of membership functions, which can be generated from ontology, looks like as follows:

**IF** $X_1$ ($\mu$ = 1/n) **THEN** $Y_1$, (1)

**IF** $X_1$ ($\mu$ = 1/n) **THEN** $Y_2$, (2)

…………………………………

**IF** $X_1$ ($\mu$ = 1/n) **THEN** $Y_n$, (n)

It should be emphasized that it does not need to generate rules in the process of membership function assignment for ontology elements in the case, when the only one ontology is available. The process of membership function assignment for ontology elements implies to assign appropriate values to the ontology elements, which are equal to rule premises, only. The process of rule generation, when there is the only one ontology, follows after the process of membership function assignment for ontology elements that is when ontology has all values of membership functions.

## 3. IMPLEMENTATION OF MEMBERSHIP FUNCTION ASSIGNMENT FOR ONTOLOGY ELEMENTS

Ontologies generally and OWL ontologies specifically are self-sufficient source to generate rules [7], [8]. Being generated rules differ in OWL ontology source code fragments, from which these rules are being obtained. Expressive means of OWL ontologies are redundant in the sense that the same information can be expressed in different ways. For example, it is possible to express a symmetric property by means of two regular properties in OWL. These redundant expressive means of OWL are good for humans, but in the same time they complicate processing of these ontologies programmatically. It is particularly sensitive to the task of membership function assignment for which properties, classes and relationships between these classes are eventually required. In this context it makes sense to normalize the OWL ontology that is to bring it to a certain standard, having only properties, classes and relationships between these classes. The process of membership function assignment to OWL ontology elements starts only after the ontology normalization process is finished.

In general all rules, which can be generated from OWL ontologies, are divided into four categories [8]:

- Identifying rules,
- Specifying rules,
- Unobvious rules or rules, generated from hidden OWL ontology elements,
- Meaning enriching rules.

Identifying rule is the rule, which determines something, based on some characteristics. Specifying rules are necessary to precise something, if this something is known. That is, specifying rules allow knowing the details of a particular object. Unobvious rules are rules, which are generated from hidden ontology elements. Hidden ontology elements are elements, which are not presented in ontology, but may be added based on the logic of ontology. Meaning enriching rules are rules that enrich existing knowledge with new details [8]. OWL ontology source code fragments, which can be transformed to unobvious and meaning enriching rules, can serve as points of application of normalization subroutines. Here normalization subroutines are understood as software components to implement the normalization process of ontology.

The ontology normalization process is divided into two separate stages. The first stage is aimed at identifying the cases, where it is possible to add unobvious elements to ontology. The second stage involves the replacement of redundant OWL elements with standard elements, where standard OWL elements are properties, classes and relations between these classes. The first stage includes the following cases of adding unobvious elements to OWL ontology (Tab. I):

TABLE I. Adding unobvious elements to OWL ontology.

| Nr | Identified pattern | Changes in pattern |
|---|---|---|
| 1 | 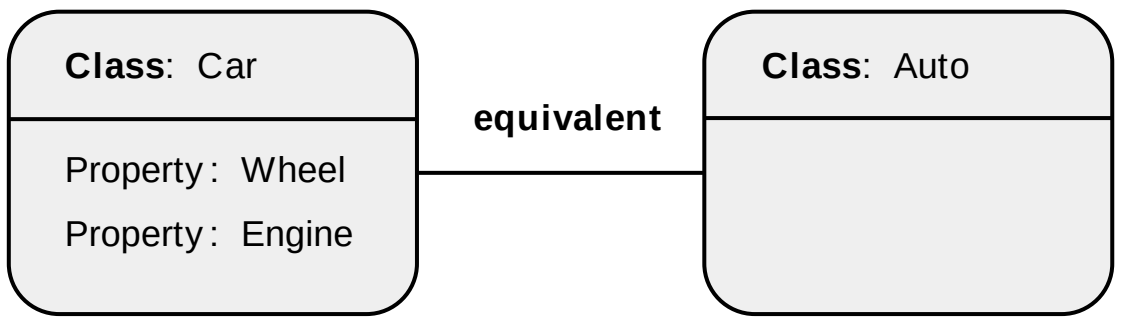  | 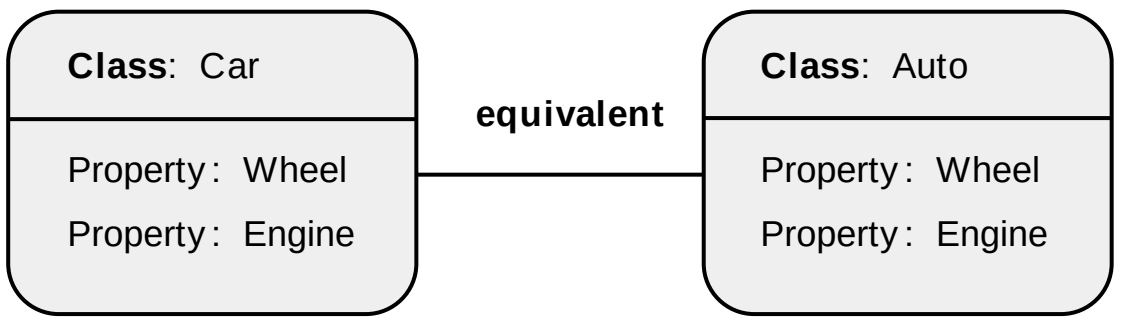  |
| 2 | 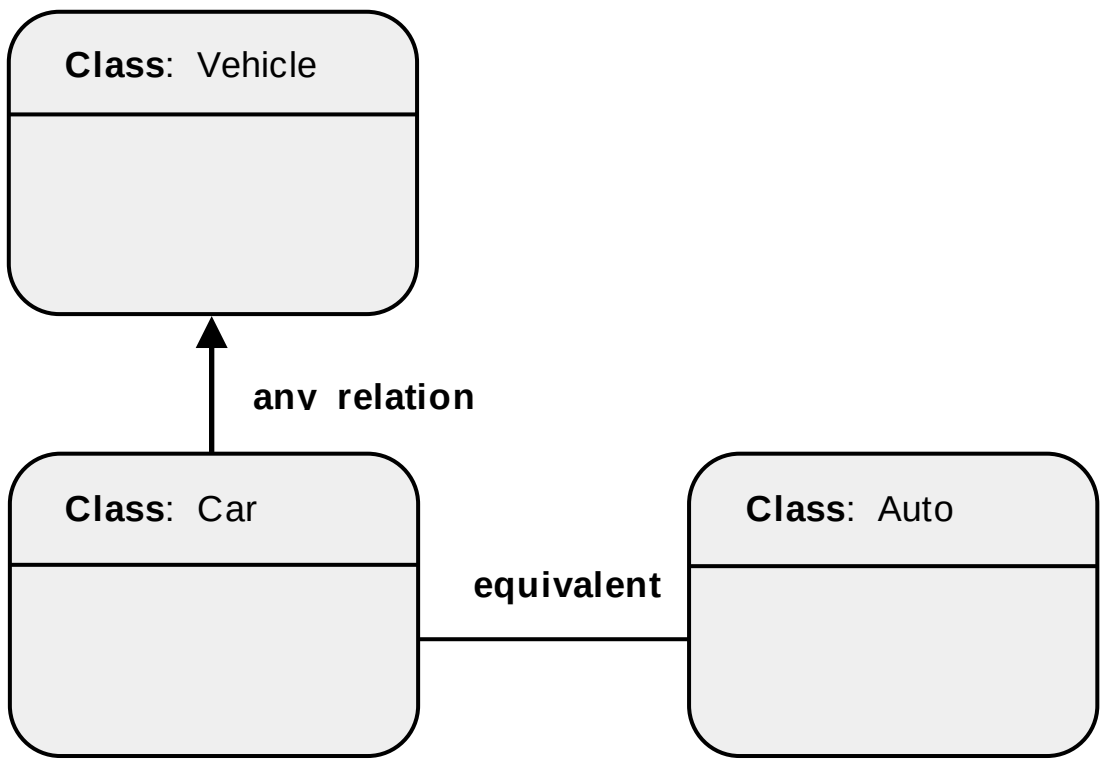  | 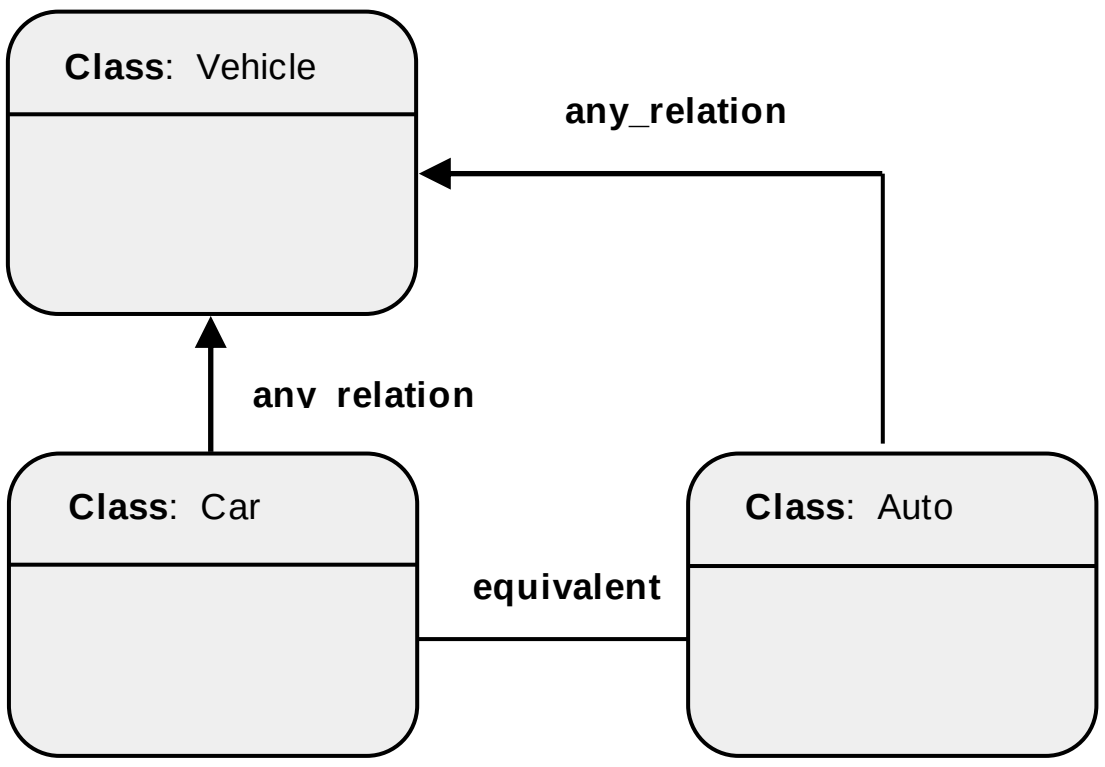  |
| 3 | 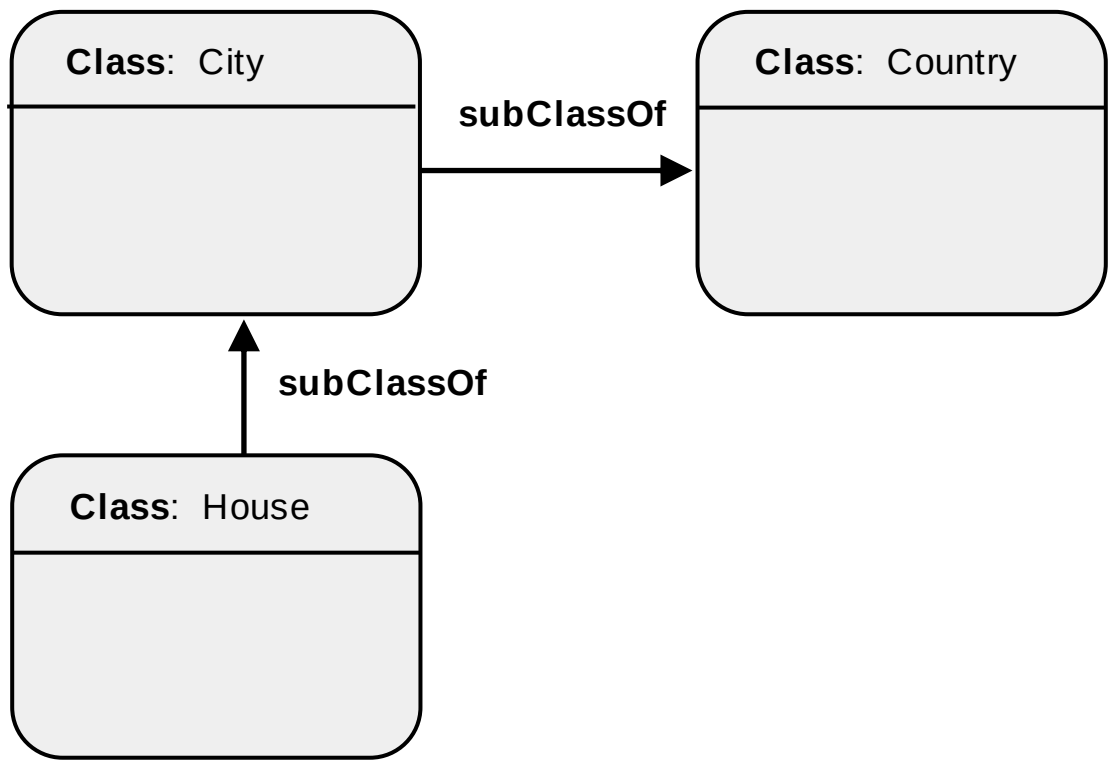 | 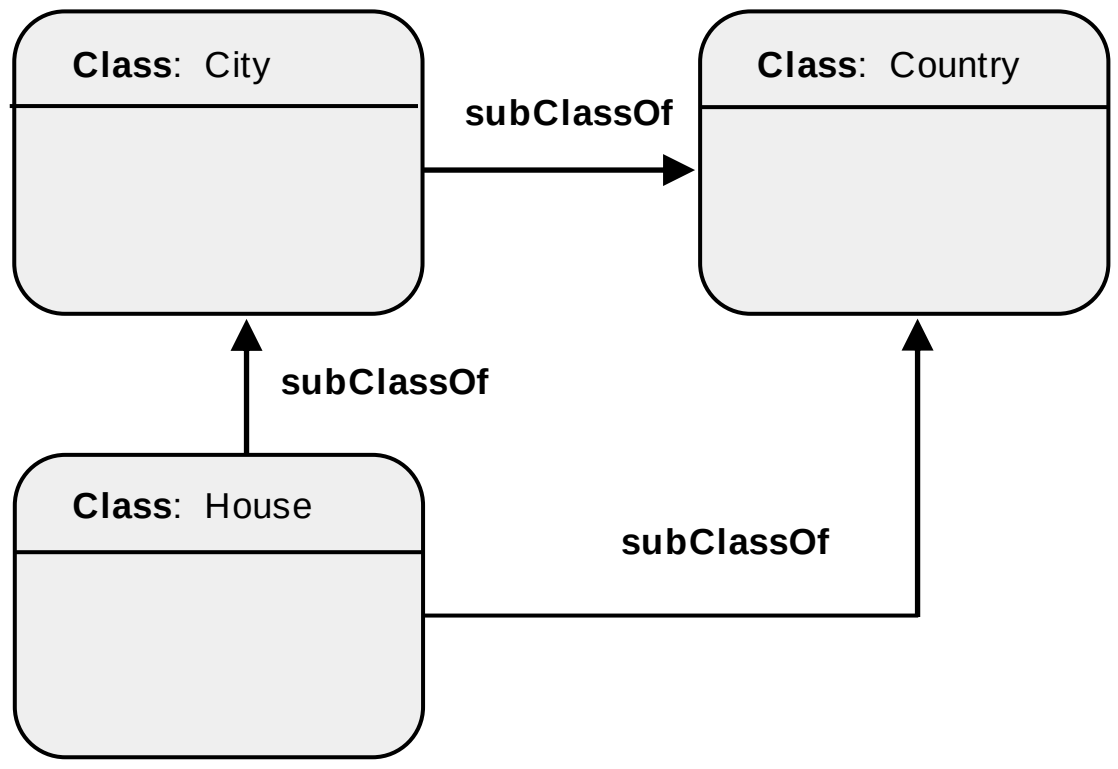 |
| 4 | 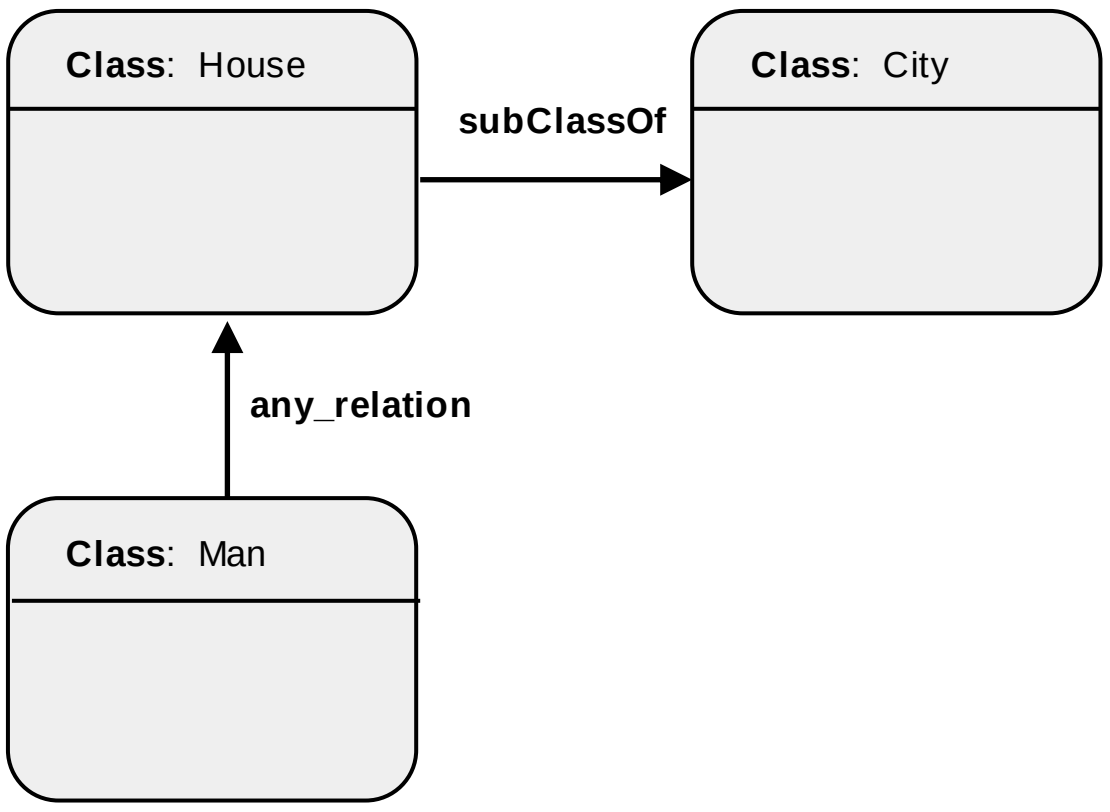 | 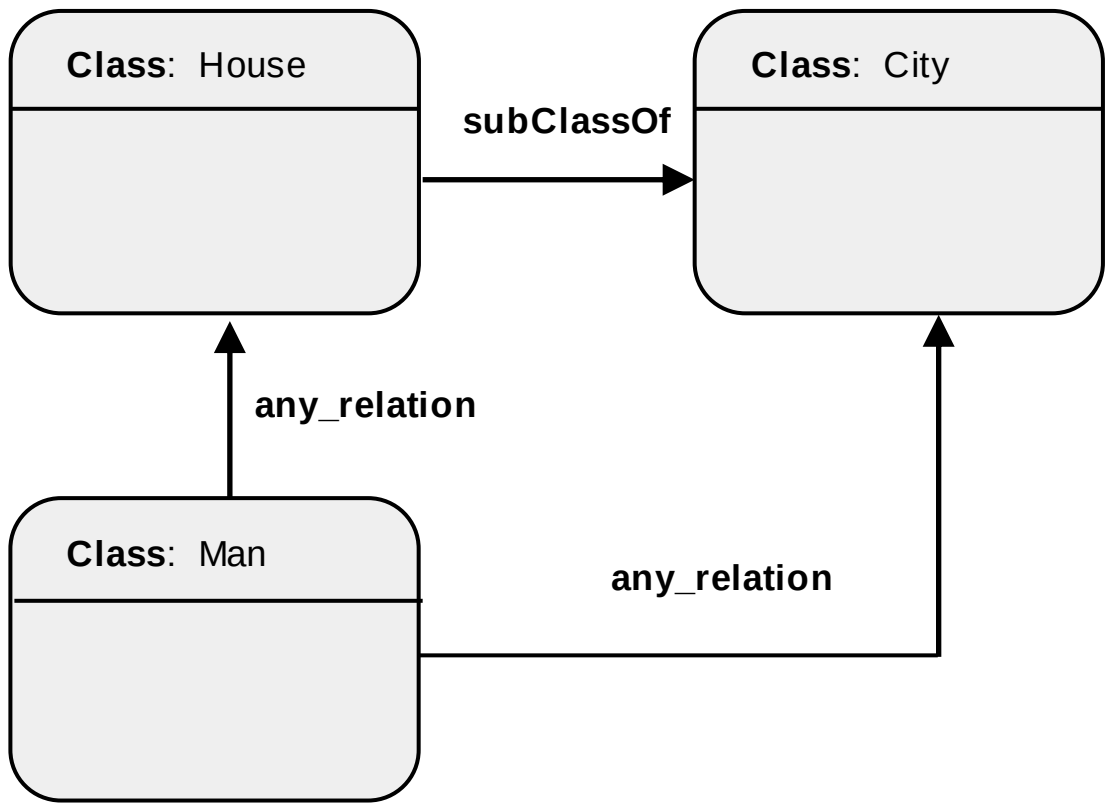 |

The first case of the table I implies adding properties from one class to another class, if these classes are equivalent. The second case implies adding relations from one class to another one, if these classes are equivalent, too. The third case is adding the unobvious "subClassOf" relation from the class "House" to the class "Country", if there are three classes "House", "City", "Country" and there are the "subClassOf" relations between the classes "House" and "City" and also between the classes "City" and "Country". The next case of the table I implies adding the relation from one class to another class (from "Man" to "City"), if there is the same relation between "Man" and "House" classes and there is "subClassOf" relation between "House" and "City" classes.

The second stage of the ontology normalization process includes the following transformations in patterns of OWL ontology source code (Tab. II):

TABLE II. Transformations in OWL ontology.

| Nr | Identified pattern | Modified pattern |
|---|---|---|
| 1 | <u>Symmetric property</u><br><br><owl:SymmetricPropertyrdf:ID="colleagueOf"><br><rdfs:domainrdf:resource="#Programmer"/><br><rdfs:rangerdf:resource="#Engineer"/><br></owl:SymmetricProperty> | <u>Object property</u><br><br><owl:ObjectPropertyrdf:ID="colleagueOf"><br><rdfs:domainrdf:resource="#Programmer"/><br><rdfs:rangerdf:resource="#Engineer"/><br></owl:ObjectProperty><br><owl:ObjectPropertyrdf:ID="colleagueOf"><br><rdfs:domainrdf:resource="#Engineer"/><br><rdfs:rangerdf:resource="#Programmer"/><br></owl:ObjectProperty> |
| 2 | <u>Inverse property</u><br><br><owl:ObjectPropertyrdf:ID="owns"><br><owl:inverseOfrdf:resource="#is_owned_by"/><br><rdfs:domainrdf:resource="#Human"/><br><rdfs:rangerdf:resource="#Plane"/><br></owl:ObjectProperty> | <u>Object property</u><br><br><owl:ObjectPropertyrdf:ID="owns"><br><rdfs:domainrdf:resource="#Human"/><br><rdfs:rangerdf:resource="#Plane"/><br></owl:ObjectProperty><br><owl:ObjectPropertyrdf:ID=" is_owned_by "><br><rdfs:domainrdf:resource="#Plane"/><br><rdfs:rangerdf:resource="#Human"/><br></owl:ObjectProperty> |
| 3 | <u>Intersection</u><br><br><owl:Classrdf:ID="Man"><br><owl:intersectionOfrdf:parseType="Collection"><br><owl:Classrdf:about="#Male"/><br><owl:Classrdf:about="#Human"/><br></owl:intersectionOf><br></owl:Class/> | <u>Subclass</u><br><br><owl:Classrdf:ID="Man"><br><rdfs:subClassOfrdf:resource="#Male"/><br><rdfs:subClassOfrdf:resource="#Human"/><br></owl:Class> |
| 4 | <u>Transitive property</u><br><br><owl:Classrdf:ID="Latgale"><br><subAreaOfrdf:resource="#Latvia"><br></owl:Class><br><owl:Classrdf:ID="EU"/><br><owl:Classrdf:ID="Latvia"><br><subAreaOfrdf:resource="#EU"/><br></owl:Class><br><owl:TransitivePropertyrdf:ID="subAreaOf"><br><rdf:typerdf:resource=http://www.w3.org/2002/07/owl#ObjectProperty/><br></owl:TransitiveProperty> | <u>Object property</u><br><br><owl:ObjectPropertyrdf:ID="subAreaOf"><br><rdfs:domainrdf:resource="#Latgale"/><br><rdfs:rangerdf:resource="#Latvia"/><br></owl:ObjectProperty><br><owl:ObjectPropertyrdf:ID="subAreaOf"><br><rdfs:domainrdf:resource="#Latvia"/><br><rdfs:rangerdf:resource="#EU"/><br></owl:ObjectProperty><br><owl:ObjectPropertyrdf:ID="subAreaOf"><br><rdfs:domainrdf:resource="#Latgale"/><br><rdfs:rangerdf:resource="#EU"/><br></owl:ObjectProperty> |

Table II presents the replacement of OWL elements with other OWL elements within the second stage of the normalization process. The first case in the table II shows the replacement of symmetric property with object property. The second one implies the replacement of inverse property with object property, too. The next case in the table II reflects the replacement of intersection element with subclass element. And the last case is aimed at the replacement of transitive property with object property.

In summary, the simplified algorithm of membership function assignment to OWL ontology elements is as follows:

1. adding unobvious elements to ontology;
2. replacing of redundant ontology elements with standard elements, where standard ontology elements are properties, classes and relationships between these classes;
3. assigning membership functions to ontology elements.

The proper process of membership function assignment to ontology elements consists of several sub processes. They are:

- membership function assignment to properties (DatatypeProperty elements);
- membership function assignment to part_of relations (subClassOf elements);

- membership function assignment to the rest relations;
- membership function assignment to the elements of equivalent classes.

The first sub process is membership function assignment to datatype properties. The formula of calculating the membership function for datatype properties is as follows (Tab. III):

TABLE III. The formula of calculating the membership function for datatype properties

| | |
|---|---|
| $$\mu_{P_i} = \frac{1}{n_{P_i}}$$ | **, where:** |
| | $\boldsymbol{i}$ – property index, |
| | $\boldsymbol{P_i}$ – property, |
| | $\boldsymbol{n_{P_i}}$ – number of classes having $P_i$ property, excluding equivalent classes, |
| | $\mu_{\boldsymbol{P_i}}$ – membership function of $P_i$ property. |

The second sub process is membership function assignment to the "part_of" (subClassOf elements) relations. The formula of calculating the membership function for the "part_of" relations is as follows (Tab. IV):

TABLE IV. The formula of calculating the membership function for "part_of" relations.

| | |
|---|---|
| $$\mu_{S_i} = \frac{1}{n_{S_i}}$$ | **, where:** |
| | $\boldsymbol{i}$ – index of (part_of + class) complex, |
| | $\boldsymbol{S_i}$ – (part_of + class) complex, |
| | $\boldsymbol{n_{S_i}}$ – number of classes, which determines complex $S_i$, excluding equivalent classes, |
| | $\mu_{\boldsymbol{S_i}}$ – membership function of complex $S_i$. |

It is necessary to make some clarifications. Complex is a single semantic object, which has single membership function. Here complex is a "part_of" relation with the resulting class.

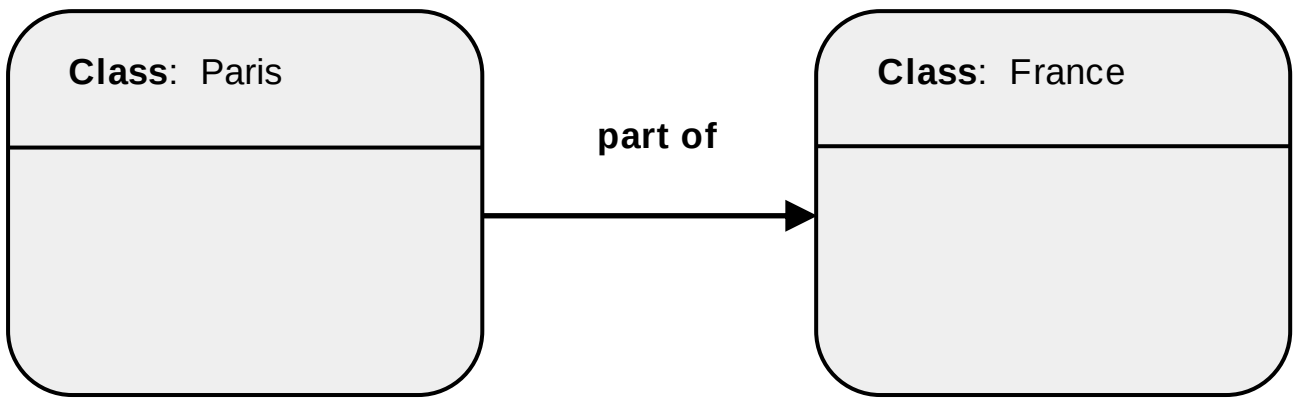

Fig.1. Example of semantic object.

In Fig.1, the semantic object "Paris part_of France" is presented. Here "part_of France" is a complex, "Paris" is a determining class and "France" is the resulting class.

The next sub process is membership function assignment to the rest relations that is relations, which are not "part_of" and equivalent. The formula of calculating the membership function for the rest relations is as follows (Tab. V):

TABLE V. The formula of calculating the membership function for the rest relations.

| | |
|---|---|
| $$\mu_{R_i} = \frac{1}{n_{R_i}}$$ | **, where:** |
| | $\boldsymbol{i}$ – index of (relation + class) complex, |
| | $\boldsymbol{R_i}$ – (relation + class) complex, |
| | $\boldsymbol{n_{R_i}}$ – number of classes, which determines complex $R_i$, excluding equivalent classes, |
| | $\mu_{\boldsymbol{R_i}}$ – membership function of complex $R_i$. |

The last sub process is membership function assignment to the elements of equivalent classes. In this case formula for calculating the membership function is not necessary. All available values of membership functions are being copied to the elements, which are associated with equivalent class: properties, "part_of" relations and other relations.

After membership functions have been assigned to OWL ontology elements, it was possible to generate fuzzy rules. Fuzzy rules are necessary for supplement of the SWES knowledge base. These rules will be utilized by semantic reasoner of the SWES to infer, based on user's data [9]. Technically the task of membership function storing for OWL ontology elements is not trivial, and it was described in previous paper in details [9].

## 4. CONCLUSION

The paper has described principles of membership function assignment for OWL ontology elements and implementation of membership function assignment for OWL ontology elements in the case, when the only one OWL ontology was available. Membership function assignment for OWL ontology elements in the case, when several OWL ontologies of one domain were available, was described in one of previous paper [4]. Implementation of membership function assignment for OWL ontology elements consisted of OWL ontology normalization process and membership function assignment process itself. The process of OWL ontology normalization has been described as a set of routines, which was focused on the OWL ontology transformation to a standard form. Here OWL ontology standard form is such OWL ontology that has certain OWL elements, only. The process of membership function assignment itself consists of four sub processes, which are necessary to assign membership functions to certain elements of OWL ontology.

Membership functions of OWL ontology elements are necessary for fuzzy rule generation, which are being stored in the SWES knowledge base and are being used by the SWES semantic reasoner to infer, utilizing user's data [9]. It is expected that users will input data, but the SWES will look for appropriate OWL ontology or ontologies in the Web and will infer, based on the data and found ontologies. Thus, the SWES will serve as such an expert system that will use knowledge from the Web.

The way of membership function assignment for the elements of the only one available OWL ontology, described in this paper, and the way of membership function assignment for the elements of several, one domain OWL ontologies, described in one of previous paper [4], are not the only ways to assign membership functions to the OWL ontology elements. There are other ways for membership function assignment for OWL ontology elements, which will be described in future papers.

Future papers will also be devoted to the further development of the Semantic Web Expert System.

## REFERENCES

[1] D. Fensel, J. A. Hendler and others "Spinning the Semantic Web," The MIT Press, Cambridge, 2005

[2] J. Davis, R. Studer, P. Warren, "Semantic Web Technologies Trends and Research in On-tology-based Systems," John Wiley & Sons Ltd, Chichester, 2006

[3] O. Verhodubs, J. Grundspeņķis, "Towards the Semantic Web Expert System," RTU Press, Riga, 2011

[4] O. Verhodubs, J. Grundspenkis, "Ontology merging in the context of a Semantic Web Expert System," Springer, Saint-Petersburg, 2013

[5] SWOOGLE. Available online: http://swoogle.umbc.edu/

[6] WATSON: Available online: http://swoogle.umbc.edu/

[7] O. Verhodubs, J. Grundspeņķis, "Evolution of Ontology Potential for Generation of Rules," Proceedings of the 2nd International Conference on Web Intelligence, Mining and Semantics, Craiova, 2012

[8] O. Verhodubs, "Ontology as a source for rule generation", 2014. Available online: http://arxiv.org/ftp/arxiv/papers/1404/1404.4785.pdf

[9] O. Verhodubs, "Adaptation of the Jena Framework for Fuzzy Reasoning in the Semantic Web Expert System", in proceedings, 2014